\documentclass[journal,twoside,web]{ieeecolor}
\usepackage{generic}
\usepackage{cite}
\usepackage{amsmath,amssymb,amsfonts}
\usepackage{algorithmic}
\usepackage{graphicx}
\usepackage{algorithm,algorithmic}
\usepackage{hyperref}
\hypersetup{hidelinks=true}
\usepackage{textcomp}
\usepackage{multirow}
\usepackage{makecell}
\usepackage{bm}
\usepackage{ulem}
\usepackage{booktabs} 
\def\BibTeX{{\rm B\kern-.05em{\sc i\kern-.025em b}\kern-.08em
    T\kern-.1667em\lower.7ex\hbox{E}\kern-.125emX}}
\begin{document}
\title{Improving Medical Visual Representation Learning with Pathological-level Cross-Modal Alignment and Correlation Exploration}
\author{Jun Wang\textsuperscript{\normalfont$\ast$}, Lixing Zhu\textsuperscript{\normalfont$\ast$}, Xiaohan Yu, Abhir Bhalerao, and Yulan He
%\thanks{%Jun Wang's PhD studentship is jointly funded by the University of Warwick and China Scholarship Council. 
%We would like to thank the University of Warwick's Scientific Computing Group for their support of this work and acknowledge the resources provided by the UKRI/EPSRC HPC platform, Sulis. }
\thanks{{$\ast$: Equal contribution.}}
\thanks{Jun Wang and Abhir Bhalerao are with the Department of Computer Science, University of Warwick, CV4 7AL Coventry, UK. (email: jun.wang.3@warwick.ac.uk, abhir.bhalerao@warwick.ac.uk)
}
\thanks{Lixing Zhu is with the Department of Informatics, King’s College London, WC2R 2LS, London, UK.}
\thanks{Xiaohan Yu is with the School of Computing, Macquarie University, NSW 2113, Sydney, Australia. Australia.}
\thanks{Yulan He is with the Department of Informatics, King’s College London, WC2R 2LS, London, UK, and the Alan Turing Institute, UK.}
}

\maketitle

\begin{abstract}
Learning medical visual representations from image-report pairs through joint learning has garnered increasing research attention due to its potential {for transferring acquired knowledge to various downstream medical tasks}. Previous works have predominantly focused on instance-wise or token-wise cross-modal alignment, often neglecting the importance of pathological-level consistency. This paper presents a novel framework PLACE that promotes the \textbf{P}athological-\textbf{L}evel \textbf{A}lignment and enriches the fine-grained details via \textbf{C}orrelation \textbf{E}xploration without additional human annotations. Specifically, we propose a novel pathological-level cross-modal alignment (PCMA) approach to maximize the consistency of pathology observations from both images and reports. To facilitate this, a Visual Pathology Observation Extractor is introduced to extract visual pathological observation representations from localized tokens. The PCMA module operates independently of any external disease annotations, enhancing the generalizability and robustness of our methods. Furthermore, we design a proxy task that enforces the model to identify correlations among image patches, thereby enriching the fine-grained details crucial for various downstream tasks. Experimental results demonstrate that our proposed framework achieves new state-of-the-art performance on multiple downstream tasks, including classification, image-to-text retrieval, semantic segmentation, object detection and {report generation.} Code is available at \href{https://github.com/Markin-Wang/PLACE}{https://github.com/Markin-Wang/PLACE}. 
\end{abstract}


\section{Introduction}
\label{sec:intro}
Powered by large-scale, high-quality data, deep learning approaches have shown significant advantages to the Computer Vision community in myriads of domains. However, in the medical domain, acquiring such an amount of finely labelled data is costly and time-consuming, let alone the inherent complexities of medical images and reports, which impedes the performance of models on various medical image processing tasks~\cite{10746601}. To mitigate this, leveraging medical reports as an auxiliary information source accompanied by the medical images in a self-supervised, joint pre-training manner has gained increasing attention. Through the aid of language information, these pre-trained models~\cite{cheng2023prior,huang2021gloria,muller2022joint,bannur2023learning} learn more generalisable image representations, which can be quickly transferred to various downstream medical tasks.

\begin{figure}
\centering
\includegraphics[width=\linewidth]{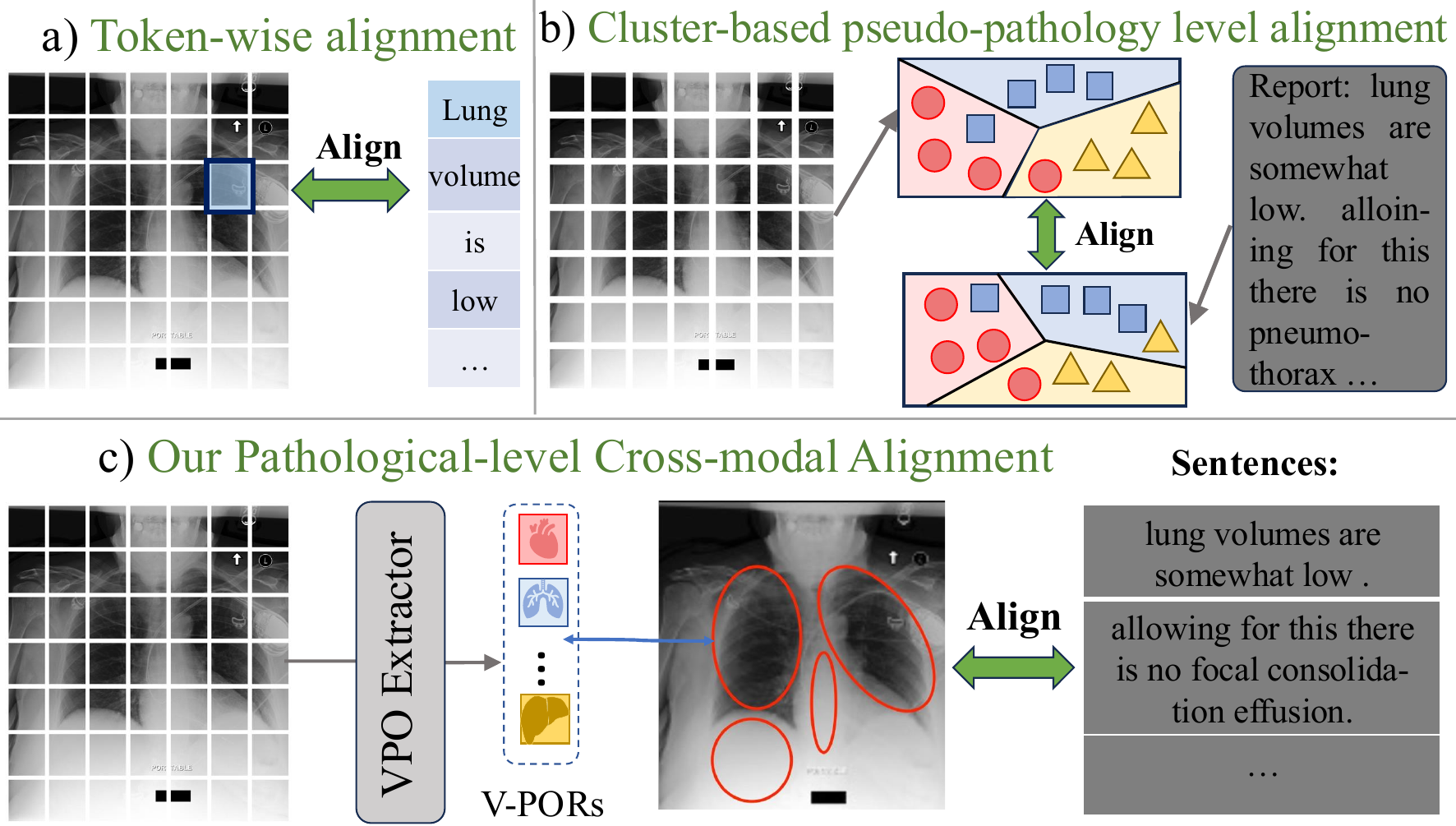}
\caption{{The illustration of (a) the token-wise alignment, (b) clustered-based pseudo-pathology level and (c) our proposed pathological-level alignment. V-PORs is the abbreviation of visual pathology observation representations which are associated with specific anatomical regions.}}
\label{fig:motivation}
\vspace{-20pt}
\end{figure}

However, directly adopting the common practice of contrastive learning for vision-language pre-training (VLP) from the usual scene to the medical domain has proved challenging. {The main reason is three-fold: medical reports usually comprise more than $4$ sentences compared to $1\sim 2$ sentences in the normal natural scene domain. This longer report covers each possible anatomical region in the image, resulting in medical image-report pairs having a more intricate cross-modal alignment and complex semantic pattern.} Additionally, different medical report pairs usually share a high similarity, while common instance-level contrastive learning pulls these pairs far apart and thus could hinder joint representation learning. Furthermore, most previous VLP works ignore the importance of the fine-grained details which play a crucial role in transferring the model to various locality-aware and disease-aware medical downstream tasks such as object detection and semantic segmentation. {The primary focus of this work is on addressing the complex cross-modal alignment issue and fine-grained representation learning.}

Some approaches have been proposed to promote the medical vision-language pre-training. For example, several studies~\cite{huang2021gloria,liu2023improving,cheng2023prior} adopt the contrastive-learning-based local alignment between image patches and words as illustrated in \autoref{fig:motivation} (a). Nonetheless, applying the local alignment on non-discriminative image patches and unimportant words introduces noisy information. Also, these token-wise alignment methods struggle to consider the pathological-level semantic information since this per-token-level granularity, e.g., one patch or word, is too small to represent pathology-level information. A group of studies~\cite{wu2023medklip,liu2023improving} improve the disease-level consistency by leveraging valuable disease labels, which however, are difficult to obtain in the real-world applications. To promote disease-{aware} alignment,~\cite{wang2022multi} proposes a disease-level cross-modal alignment~\cite{wang2022multi}.~\cite{li2024mlip} further extend~\cite{wang2022multi} by further distilling knowledge information into the disease-level cross-modal alignment. Although improvements have been demonstrated, these so-called disease-level alignments are conducted on clustered representations as depicted in \autoref{fig:motivation} (b). Hence, they are less effectively applied to pathology alignment and might not adhere closely to a pathological-level alignment.

To this end, we propose a novel pathological-level cross-modal alignment (PCMA) to help learn a pathology enriched, more generalizable image representation where \autoref{fig:motivation} (c) shows a high-level illustration. Specifically, our PCMA module learns multiple pathological observation representations (PORs) from {each image-report pair}, and performs the alignment between the PORs {within each sample} without extra disease label annotations. We further design a visual pathology observation extractor (VPOE) which leverages the learnable pathological query tokens with a transformer to automatically retrieve visual PORs from high-level visual representations, while the textual PORs are derived from the sentence-based textual tokens. To enable the model to capture more fine-grained details and promote the cross-modal representation learning, we further introduce a cross-modal correlation exploration task which enforces the model to recognize the correlation among different image patches purely conditioned on reports. {This objective compels the model to capture correlations among diverse image patches from the report, enhancing the model's comprehension of intricate semantic patterns and discourse correlations. Ultimately, this improves the model's ability to learn a locality-aware and fine-grained representation, which holds significance for downstream tasks such as object detection and semantic segmentation.} Our proposed framework ensures both the instance-level and pathological-level alignments with fine-grained details enrichment on medical image-report pairs. {Note that disease observations typically pertain to abnormal clinical conditions defined by specific symptoms and causes, while pathological observations encompass both normal and abnormal findings.}

Our work has three principal contributions:
\begin{enumerate}
    \item We propose a novel pathological-level cross-modal alignment module by maximizing the mutual information between visual and textual pathological observations {within each sample}. Compared with previous work, our PCMA is more robust, pathologically generalizable and effectively applied without reliance on extra disease-level labels.
    \item We present a cross-modal correlation exploration task requiring the model to predict the correlation among different image patches from the reports. Such a correlation prediction task guides the model to learn a more fine-grained image representation while simultaneously enhancing the cross-modal representation learning. 
    \item We demonstrate our framework to yield new state-of-the-art performance on a variety of downstream tasks: medical classification, object detection, semantic segmentation, and image-to-text retrieval.
\end{enumerate}

\begin{figure*}[t]
\centering
\includegraphics[width=0.95\textwidth]{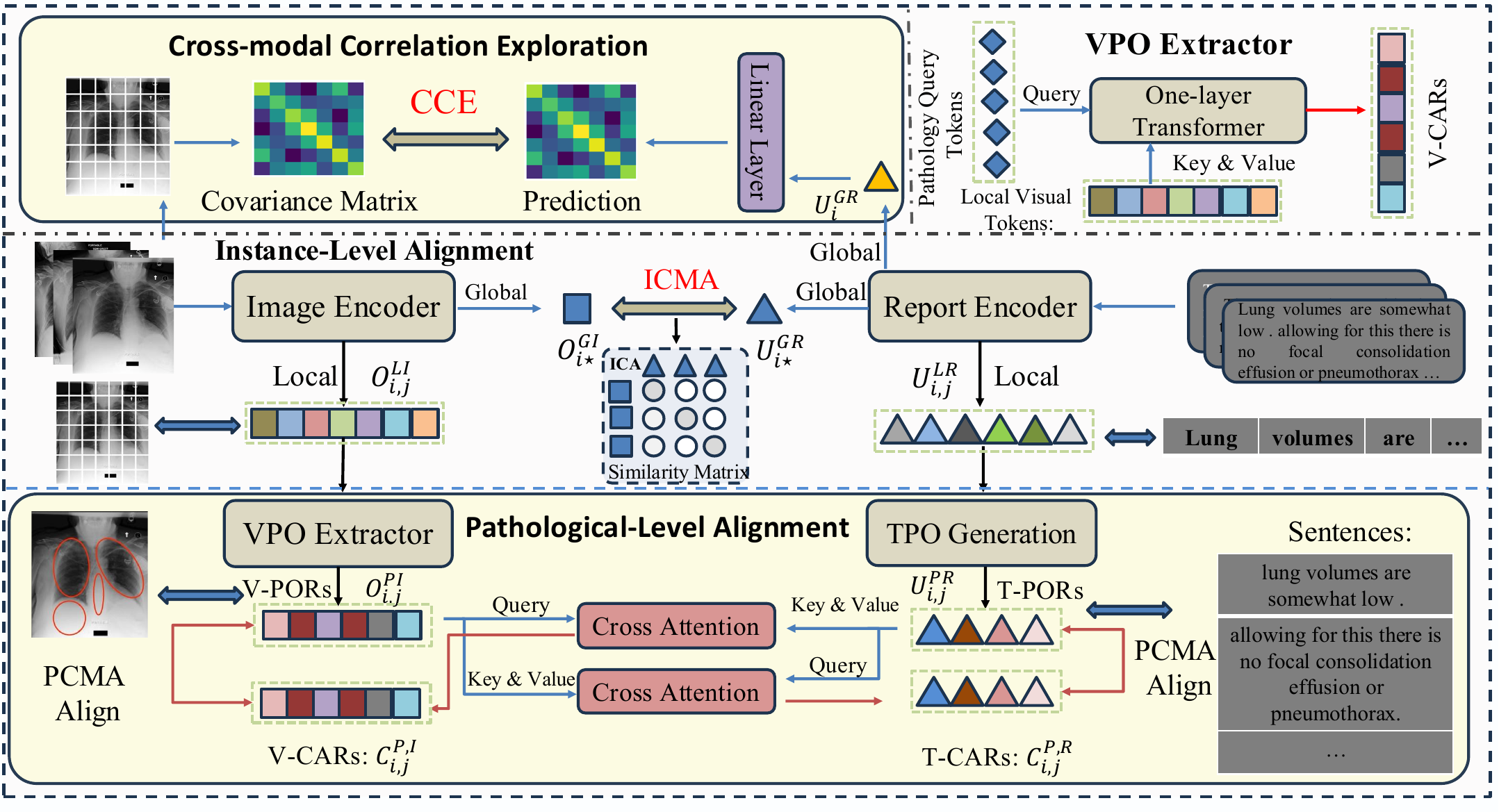}
\caption{{The overall architecture of $PLACE$. Our model takes advantage of both the instance-level and pathological-level cross-modal alignments during the joint-training. The proposed Visual Pathology Observation Extractor utilizes the learnable pathology query tokens to derive the visual pathology observation representations which are then aligned with the textual pathology observation representations by the PCMA module. The cross-modal correlation exploration module calculates the covariance matrix for the image patches and requires the model to predict this matrix based on the global report representation. Through these carefully designed objectives, PLACE is capable of learning a more generalizable and fine-grained visual representation.} }
\label{fig:architecture}
\vspace{-20pt}
\end{figure*}

\section{Related Work}
\label{sec:related_works}
\subsection{Medical Image and Report Pre-training}
\label{sec:medvlp}
In recent years, great effort~\cite{li2023blip,liu2024improved,wang2024visionllm,huang2023language,boecking2022making,radford2021learning,zhuminigpt,alayrac2022flamingo} has been made to explore image-text joint training. By enforcing a cross-modal alignment, the learnt representation shows a potent generalisation capability that can be transferred to various multimodal tasks. This image-text joint training also serves to enhance visual representations with the incorporation of text, thereby playing a pivotal role in the medical domain where having a large-scale labelled dataset to train a single model is not only labour-intensive but also requires specialized expertise. 

Numerous methods%~\cite{bannur2023learning,moon2022multi,wan2024med,zhang2023knowledge,wu2023medklip,liu2023m,faghri2018vse++,engilberge2018finding} 
~\cite{bannur2023learning,moon2022multi,wan2024med,zhang2023knowledge,wu2023medklip,liu2023m} have been devised to advance medical image-report joint training. These approaches mainly follow a contrastive learning based cross-modal alignment paradigm. The initial attempt~\cite{zhang2022contrastive} follows CLIP~\cite{radford2021learning} to pre-train the model from paired images and reports. Nonetheless, the global-level contrastive learning in vanilla CLIP ignores the high similarities among different medical image-report pairs.~\cite{wu2023medklip,liu2023improving,liu2023m} mitigates this problem by introducing similarity-based soft labels as the matching targets. Although improvements have been seen, they require the disease labels to calculate the similarities, which, however, can be difficult to obtain in a medical domain, limiting its generalization capability. Another group of methods~\cite{wang2022multi,cheng2023prior,huang2021gloria} cope with this problem through some localized-based contrastive alignment methods, e.g., image patch$\leftrightarrow$word alignment. Some studies~\cite{wu2023medklip,zhang2023knowledge,cao2024bootstrapping} leverage extra data sources, e.g. disease labels, knowledge graphs, Unified Medical Language System (UMLS)~\cite{lindberg1993unified} or even professional expertise to enrich medical knowledge during image-report joint pre-training. These approaches ignore the importance of pathological-level alignments which serves as a critical component when transferring a learnt representation to various medical downstream tasks, or require valuable disease labels. There are a few studies~\cite{wang2022multi, li2024mlip} that investigate the pathological-level alignment without disease labels by constructing a cross-modal alignment between the clustered image and the report representations. However, they are less effectively applied to pathological information and do not closely conform to pathological-level alignment since clustered representations normally struggle to capture pathologies in long-text scenarios. Furthermore, most previous studies overlook the significance of fine-grained details, such as lesions, which are essential for medical localization downstream tasks. Some approaches%~\cite{chen2023towards,huh2023improving,chen2022multi,li2024llava,chen2022align,wang2024camanet,lin2023pmc}
~\cite{chen2023towards,huh2023improving,li2024llava,wang2024camanet} aim to address multimodal tasks while this work focuses more on learning a useful medical visual representation which may be usefully applied to a variety of visual-based downstream tasks.

To this end, we propose a novel framework PLACE that ensures the cross-modal alignments at both instance and pathological level without extra human annotations being required. Additionally, we develop a proxy task that compels the model to ascertain the correlation among image patches through the reports, thereby facilitating the capture of more fine-grained details and further enhancing the cross-modal alignment.

% \subsection{Self-supervised Training}
% \label{sec:sst}

\section{Method}
\label{sec:method}
This section presents the details of our proposed methods. Typically, the vision-language joint learning task endeavours to acquire a joint distribution $P(X,R)$ between the medical image set $X=\{x_1, x_2...,x_N\}$ and the corresponding report set $R=\{r_1,r_2,...,r_N\}$. Each sample $e_i =<x_i,r_i>$ in our context constitutes a medical image-report pair. With minimal training or a zero-shot scenario, a well-developed medical visual representation can be effectively adapted to various downstream tasks and exhibit favourable performance.

\subsection{Framework Overview}
\label{sec:framework}
Our proposed approach enriches fine-grained details and improves pathological-level alignment, so as to facilitate the acquisition of a more efficacious joint representation. As depicted in~\autoref{fig:architecture}, our framework comprises three principal components: 1) an instance-level cross-modal alignment; 2) a pathological-level cross-modal alignment; 3) a cross-modality correlation exploration proxy-task to ensure that the model captures fine-grained details while augmenting cross-modal representation learning. The details of each component are elaborated in the subsequent sections.

\subsection{Image and Report Representation}
\label{sec:ir_encdoe}
Given an image-report pair $<x_i,r_i>$, the first step is to obtain the representations for both the image and the report. Specifically, an image encoder $f_I$, e.g., ResNet50~\cite{he2016deep} or ViT~\cite{dosovitskiy2021an}, transforms the image $x_i$ into a subregion representation, which is then flattened to a sequence of local visual tokens formed as $O^{LI}_{i,j}$. $i$ and $j$ denote $i^{th}$ sample in the batch and the $j^{th}$ patch in the image. Thereafter, average pooling is applied over the local visual tokens to derive the global image representation $O^{GI}_{i}$. $L$ and $G$ designate the global or local representation. Similarly, the report $r_i$ is embedded and encoded into a sequence of local textual tokens $U^{LR}_{i,j}$ by a text encoder $f_R$. $j$ refers to the $j^{th}$ token in the report. We follow the common practice of considering the $[CLS]$ token as the global report representation denoted by $U^{GR}_{i}$. 

\subsection{Instance-level Cross-modal Alignment}
\label{sec:ica}
Images and their corresponding reports typically exhibit a substantial semantic congruity. The instance-level cross-modal alignment (ICMA) stage seeks to draw paired samples together while distancing the unpaired samples within the latent space at a global image-report level. This is accomplished by maximizing the mutual information between the image-report pairs, as articulated by \autoref{eq:mutual_info}. 
\begin{align}
\label{eq:mutual_info}
  \bm{\mathcal{I}}(\bm{\mathit{X}},\bm{\mathit{R}}) = \sum_{x,r}\mathit{p}(x,r)\log\frac{p(x|r)}{p(x)}.
\end{align} 
In particular, we first map the global image and report representation into a latent space $D$ through two $MLP$ layers $m_I$ and $m_R$. After that, the symmetric InfoNCE loss~\cite{oord2018representation} is adopted to optimise the lower band of \autoref{eq:mutual_info}, thus enhancing the mutual information. \autoref{eq:r_to_i} and \autoref{eq:i_to_r} illustrate the process of this cross-modal alignment at the instance level:
\begin{align}
\label{eq:r_to_i}
  \mathcal{L}_{ICMA}^{I\gets R} &= -\frac{1}{B} \sum_{i=1}^B\log \frac{\exp(O^{GI}_{i\star} \cdot {U^{GR}_{i\star}}^T / \tau_1)}{\sum_{j=1}^B \exp(O^{GI}_{i\star} \cdot {U^{GR}_{j\star}}^T  / \tau_1)}, \\
\label{eq:i_to_r}
  \mathcal{L}_{ICMA}^{R\gets I} &= -\frac{1}{B} \sum_{i=1}^B\log \frac{\exp(U^{GR}_{i\star} \cdot {O^{GI}_{i\star}}^T / \tau_1)}{\sum_{j=1}^B \exp(U^{GR}_{i\star} \cdot {O^{GI}_{j\star}}^T  / \tau_1)}.
\end{align} 

\noindent Here $O^{GI}_{i\star}$ and $U^{GR}_{i\star}$ are the transformed global image and report representation obtained via $O^{GI}_{i\star} = m_I(O^{GI}_{i})$ and $U^{GR}_{i\star} = m_R(U^{GR}_{i})$, respectively. $\tau_1$ refers to the softmax temperature.

\subsection{Pathological-level Cross-modal Alignment}
\label{sec:pcma}
An objective of cross-modal alignment at the instance level has demonstrated effectiveness in the acquisition of joint representations within the domain of natural scenes. However, as a characteristic of medical VLP, distinct image-report pairs can demonstrate significant semantic similarity due to the subtle differences among images and the high similarities among reports. {Models trained solely on instance-level contrastive loss struggle to learn a meaningful representation from samples sharing high similarities such as similar images and reports.} Furthermore, various downstream medical tasks are increasingly reliant on a clinically accurate representation.

%Solely applying instance-level cross-modal alignment (ICMA) inadvertently separates these high-similarity pairs, which could lead to ill-posed outcomes and potentially impede the learning process. Furthermore, various downstream medical tasks are increasingly reliant on a clinically accurate representation.

To this end, we propose a novel Pathological-level Cross-modal Alignment (PCMA) module that promotes consistency among image-report pairs in pathological observations. %Our PCMA predominantly focuses on the subject level, hence ameliorating the limitations previously noted in the ICMA, while rendering the model capturing more pathological information. 
{The PCMA module operates at a finer subject-level within each sample. Specifically, it is developed to bring the anatomical region representation closer to its associated textual representation while creating distance between unpaired anatomical regions and sentences. This contrastive learning process is carried out with each sample. Consequently, the PCMA module is less likely to be affected by high similarities among different samples, and encourages the model to delve deeper into the pathological structure within the samples.}

To achieve this, we must obtain the pathological observation representation (POR) for both the image and the report. In particular, each sentence within the report is posited to correspond to an anatomical region depicted in the image, articulating specific observations thereof. Therefore, we adopt a straightforward way to construct the Textual Pathological Observation Representation (T-POR) by computing the mean of the representations of tokens within each sentence. {Specifically, a full report $r$ consists of several sentences denoted as $r=\{sent_1,sent_2,...,sent_{N_s}\}$, The $k$-th T-POR in the $i$-th report $U^{PR}_{i,k}$ is calculated as:
\begin{equation}
    U^{PR}_{i,k} = \frac{1}{N_t} \sum_{j\in s_k} U^{LR}_{i,j}
\end{equation}
\noindent where $U^{LR}_{i,j}$ is the $j$-th textual token representation in $k$-th sentence obtained from the last layer of the text encoder. $N_s$ refers to the number of sentence in the report.}

After having the \textit{textual} pathological observation representation on hand, the next step is to acquire the \textit{visual} pathological observation representation (V-POR). However, unlike the report, deriving the V-POR without accurate bounding box annotations and disease labels would be of great difficulty. Motivated by~\cite{li2023blip} which demonstrates that images of any resolution can be transformed to a fixed set of visual tokens while maintaining the semantic information, we propose to extract a V-POR by exploiting the intrinsic structure of the image-report pair and by the aid of the T-POR without extra human annotations. Specifically, we first introduce a visual pathology observation extractor (VPOE) which is a transformer-based architecture ({adhering to a Self-Attention $\rightarrow$ Cross-Attention structure}) with a sequence of learnable pathology query tokens $Q\in \mathbb{R}^{N_q \times D}$. $N_q$ denotes the number of query tokens. %and each learnable query token is functioned to retrieve one type of visual pathological observation from the visual tokens. Then, given the local visual tokens $U^{LI}_i$ and query tokens $Q$, the process to obtain the V-POR for $l^{th}$ transformer layer $O^{PI}_{i,l}$ in VDE can be summarize as\footnote{we omit the $\mathrm{FeedForward}$ and $\mathrm{Add\&Norm}$ after each attention layer for clarity.},:
{Query tokens, designed to extract specific V-PORs from localized visual tokens, first performs a self-attention mechanism to capture contextual relationships within themselves. Then, they interact with these localized visual tokens through a cross-attention mechanism within the VPOE to align/extract the VPOs. Here, the query tokens serve as queries, while the localized visual tokens act as keys and values. The proposed PCMA module utilizes T-POR and contrastive learning to guide the model in exploiting the inherent structure of the image-report pair. This supervision aids in training the pathology query tokens to extract the most pertinent V-POR. Given the local visual tokens $U^{LI}_i$ and query tokens $Q$, the process to obtain the V-POR for $l^{th}$ transformer layer $O^{PI}_{i,l}$ in VDE is summarized by~\autoref{eq:vde}. The upper right part of~\autoref{fig:architecture} also illustrates this process.}
\begin{align}
\label{eq:vde}
  O^{PI}_{i,l} &= \mathrm{CrossAttn}(\mathrm{SelfAttn}(Q_{l-1}), U^{LI}_i)),
\end{align} 
where $\mathrm{-Attn}$ refers to a vanilla attention mechanism and $Q_0 = Q$. We consider the output of the last transformer layer in $VDE$ as the final visual pathological observation representation denoted by $O^{PI}_i$. After that, we enforce the alignment between the visual and textual POR. Nevertheless, there are no ground truth matching annotations between the visual and textual POR since the T-POR for one anatomical region can occur at any position in the report. Similarly to~\cite{cheng2023prior,huang2021gloria}, we adopt a cross-attention mechanism to investigate the matching between visual and textual PORs. Specifically, for the T-POR $O^{PI}_{i,j}$, we generate its corresponding cross-modal attended representation $C^{PI}_{i,j}$ via:
\begin{align}
\label{eq:cmar_i1}
  a^{j,k}_{i} &= \mathrm{softmax}(\frac{O^{PI}_{i,j} \cdot U^{PR}_{i,k})}{\sqrt{D}}), \\
\label{eq:cmar_i2}
  C^{PI}_{i,j} &= \sum\nolimits_{k=1}^{N_s} a^{j,k}_{i} \cdot U^{PR}_{i,k}.
\end{align} 
Then, similar to ICMA, we apply the symmetric InfoNCE loss to maximise mutual information between a V-POR $O^{PI}_{i}$ and its corresponding visual cross-modal attended representations (V-CARs) $C^{PI}_{i}$. Pathological-level report-to-image alignment loss $\mathcal{L}^{I\gets R}_{PCMA}$ is formulated as:
\begin{align}
\label{eq:pca}
\begin{split}
  \mathcal{L}^{I\gets R}_{PCMA} &=-\frac{1}{2BN_v} \sum_{i,j}\log \frac{\exp(O^{PI}_{i,j} \cdot {C^{PI}_{i,j}}^T / \tau_2)}{\sum_{k=1}^{N_q} \exp(O^{PI}_{i,j} \cdot {C^{PI}_{i,k}}^T  / \tau_2)}, \\
  &+\log \frac{\exp(U^{PI}_{i,j} \cdot {O^{PI}_{i,j}}^T / \tau_2)}{\sum_{k=1}^{N_q} \exp(U^{PI}_{i,j} \cdot {O^{PI}_{i,k}}^T  / \tau_2)}.
\end{split}
\end{align} 
The pathological-level report-to-image alignment loss $\mathcal{L}^{R\gets I}_{PCMA}$ can be acquired in a similar way. We observe that the importance of different sentences (Textual Pathological Observations) obviously varies. For example, sentences presenting pathologies or abnormal observations play a more crucial role in PCMA. Hence, we further add a weight for each T-POR when calculating the $\mathcal{L}^{R\gets I}_{PCMA}$. This weight is calculated by aggregating the attention scores over all tokens in the sentence to the $[CLS]$ token. {Specifically, assuming the attention score of $j$-th textual token in $i$-th sentence is $s_{i,j}$, the weight for $i$-th T-POR $w'_{i}$ is calculated as:} \\

{
\begin{equation}
    \label{eq:weight_cal}
w'_i = \frac{w_i}{\sum_k{w_k}}, \quad \quad w_{i} = \sum_{j\in sent_i}{a_j}.
\end{equation}
}

\noindent {A normalization is applied to map the initial weight score $w_i$ into $[0,1]$.} The proposed PCMA module enforces the VDE to retrieve the V-POR most relevant to the textual pathological observations, and the well-learnt V-POR will further improve the PCMA performance in return, showing a characteristic of online refinement and complementary module.

% \noindent \textbf{Discussion.} We employ a lightweight Q-Former style ~\cite{li2023blip} architecture to implement the VPOE generating V-PORs which has never been seen in previous works. The Q-Former-like architecture is a kind of architecture that has been adopted in several multimodal studies~\cite{zhang2023video,dai2023instructblip,chen2023x} which normally focus on developing retrieved tokens interpretable by large language models through objectives such as instance-level image-text matching and text generation. However, the function of the proposed PCMA and learning process for the retrieved tokens diverge significantly from them. We anticipate that each retrieved token produced by the VPOE encapsulates the representation of a single type of visual pathology observation, which is subsequently aligned with textual pathology observations to facilitate alignment at the pathological level. To achieve this aim, we devise the PCMA learning objective to guide the learning of retrieved tokens and bridge the modality gap at a more fine-grained, pathological level. The Q-former style architecture serves as a method to generate V-PORs here, while the learning objective and the underlying rationale, i.e. the pathological-level alignment and leveraging query tokens to generate V-PORs, assume a position of greater significance.

\subsection{Cross-modal Correlation Exploration}
\label{sec:CCP}
Various medical downstream tasks, e.g. detection and segmentation, require a learnt representation containing more fine-grained details, e.g. low-level visual information about lesions, to better capture subtle differences among different samples. Masked image modeling (MIM)~\cite{xie2022simmim}, a widely utilized self-supervised method aimed at enhancing the learning of fine-grained details through the prediction of raw pixels in masked regions, has demonstrated efficacy within the natural scene domain. Nonetheless, its application within the medical domain encounters challenges, as the masked image regions can potentially disrupt the semantic continuity to which radiological interpretation is highly sensitive~\cite{chen2022multi}, while implementing a separate forward process for the MIM objective considerably increases the training complexity. Moreover, conventional MIM is typically conducted within a single-modal scenario, wherein the pixel prediction of the masked areas relies solely on visual representation, constraining its effectiveness in cross-modal applications.

To this end, we therefore propose a novel Cross-modal Correlation Exploration (CCE) task to help the model capture more fine-grained details while simultaneously improving the cross-modal understanding without breaking the semantic continuity. The covariance reveals the correlation among two variables. After observing that medical images normally show a correlation on both the relative positions and contents among different anatomical regions, we evenly split the image into $N_p$ patches with a patch size of $PS$ and regard the raw pixel of each patch as a variable. Denoting the raw pixel of $i^{th}$ patch in the image as $e^i$, the covariance $cov(e^i,e^j)$ indicates the correlation between $i^{th}$ and $j^{th}$ patches calculated by:
\begin{align}
\label{eq:cov_cal}
cov(e^i,e^j) = \sum\nolimits_{k=1}^{N_c}\frac{(e^i_k - \bar{e}^i) \cdot (e^j_k - \bar{e}^j)}{N_c - 1},
\end{align} 
\noindent where $\bar{e}^i$ and $\bar{e}^j$ are the mean of $e^i$ and $e^j$ respectively. By calculating the covariance among each pair of patches (variables), we obtain a covariance matrix $\bm{\Sigma}^i \in\mathbb{R}^{N_p\times N_p}$ describing the correlation among different image patches in the $i^{th}$ sample. After that, to enforce that the model capture further fine-grain details and promote the cross-modal representation learning, we design a novel proxy task which requires it to predict this covariance matrix conditioned on the global report representation. We simply employ one linear layer as the CCP head to predict the covariance matrix: $\bm{H}^i = W_h \cdot \bm{U}^{GR}_i$ where $\bm{H}^i$ refers to the predicted covariance matrix for $i^{th}$ sample. After that, the Mean Squared Error (MSE) loss is used to supervise the learning of the cross-modal correlation prediction:
\begin{align}
\label{eq:ccp}
\mathcal{L}_{CCE} = \frac{1}{B}\sum_{i=1}^{B}(\frac{1}{{N_p}^2}\sum_{j=1}^{N_p}\sum_{k=1}^{N_p}{(\bm{\Sigma}^i_{j,k} - H^i_{j,k})}^2)
\end{align} 

The proposed objective CCE necessitates a significantly more fine-grained details understanding compared to the traditional MIM since it demands not only that the model comprehends the content within each image region from reports but also discerns the potential correlations among various image regions. {The CCE objective uses the global report representation to predict the covariance matrix. Gradients from the CCE head backpropagate through the textual encoder. The global textual representation interacts with localized textual tokens via self-attention, and these localized textual tokens further interact with V-PORs through cross-attention, as specified in the PCMA module. As a result, both the VPO Extractor and the visual encoder—which generates the localized visual tokens—are also influenced by the CCE objective.}
\begin{table*}
	\setlength{\abovedisplayskip}{-0.0cm}
	\centering
	\resizebox{0.9\linewidth}{!}{
		\begin{tabular}{ccccccc|cccccc}
			\toprule
			\multirow{2}*{Method} & \multicolumn{3}{c}{RSNA (Dice)} & \multicolumn{3}{c|}{SIIM (Dice)} & CheXpert & \multicolumn{4}{c}{COVIDx (ACC)}  \\
			& 1\% & 10\% & 100\% & 1\% & 10\% & 100\% & 0-shot & 0-shot & 1\% & 10\% & 100\% \\
			
			\hline
			%Random Init & 6.9 & 10.6 & 18.5 & 	9.0 &  28.6 & 54.3 & - & - & 50.5 & 60.3 & 70.0 \\
			ImageNet Init & 34.8 & 39.9 & 	64.0&  10.2 & 35.5 & 63.5 &- &- & 64.8 & 78.8 & 86.3 \\
			ConVIRT \cite{zhang2022contrastive}& 55.0 & 67.4 & 67.5 & 	25.0&  43.2 & 59.9 & 47.6 & 17.8 & 72.5 & 82.5 & {92.0}\\
             GLoRIA-CheXpert \cite{huang2021gloria}& 59.3 & 67.5 & 67.8 & 	35.8& 46.9 & 63.4 & 50.4 & 20.9 & 67.3 & 77.8 & 89.0\\
			GLoRIA-MIMIC \cite{huang2021gloria}& 60.8 & 68.2 & 67.6 & 37.6& 56.4 & 64.0 & 51.7& 22.1 & 67.3 & 81.5 & 88.6\\
			MGCA (ResNet-50) \cite{wang2022multi}&63.0 & 68.3 &69.8 & 49.7 & 	59.3&  64.2 & 50.2 & 24.5 & 72.0 & 83.5 & 90.5\\
            MedKLIP (ResNet-50) \cite{wu2023medklip}&66.2& 69.4 & 71.9 & 	50.2&  60.8 & 64.4 & - & - & {74.5} & {85.2} & 90.3\\
            PRIOR (ResNet-50) \cite{cheng2023prior}&66.4  & 68.3 & 72.7 & 51.2&  59.7 & 66.3 & 56.3& 25.9 & 72.3 & 84.7 & 91.0\\
       M-FLAG (ResNet-50) \cite{liu2023m}&64.6  & {69.7} & 70.5 & {52.5}&  {61.2} & 64.8 & 55.9& 25.4 & 72.2 & 84.1 & 90.7\\
            MLIP ( ResNet-50) \cite{li2024mlip}& {67.7}  & 68.8 & {73.5} & 	51.6& 60.8 & {68.1} & {56.9} & {26.3}& {73.0}& {85.0} &90.8\\
        ASG ( ResNet-50) \cite{li2024anatomical}& {68.4}  & 69.9 & {72.6} & 60.7& {66.7} & {72.6} & - & -& -& - &93.3\\
        G2D ( ResNet-50) \cite{liu2024g2d}& {70.9}  & {72.6} & {75.1} & {62.6}& {63.1} & {66.8} & - & -& {76.6}& {88.2} &{93.3}\\
          \textbf{PLACE} (Ours, ResNet-50)& \textbf{74.2}  & \textbf{76.4}  &  \textbf{77.0} & \textbf{64.7} 	& \textbf{73.5}  &\textbf{73.8}    &\textbf{63.5} & \textbf{44.0}& \textbf{76.8}& \textbf{89.3} &\textbf{94.0} \\
        \midrule
            
			MGCA (ViT-B/16) \cite{wang2022multi}& - & - & -&	-&-  & -& {50.0}& {33.2} & 74.8& 84.8 & {92.3}\\
			MLIP (ViT-B/16) \cite{li2024mlip}& - & -& - & 	-&  -&  - &  {57.0}& {34.8} & {75.3}&{86.3}& {92.5}\\
			%			MLIP(Ours, ViT-B/16)
          
            \textbf{PLACE} (Ours, ViT-B/16) & -  &-  & - & -& - &-  & \textbf{61.8} &\textbf{41.7} & \textbf{77.5}& \textbf{90.0} &\textbf{93.3}\\
        \bottomrule
	\end{tabular}}
	\caption{Results of semantic segmentation and image classification in a zero-shot setting (for classification), 1\%, 10\% and 100\% training samples. The evaluation metric for CheXpert is AUC. The best results are highlighted in bold.}
	\label{tab:seg_and_cls}
\vspace{-15pt}
\end{table*}

\subsection{Overall Objective}
\label{sec:overall_obj}
Our proposed model is jointly trained with the three modules, i.e. the ICMA, PCMA and CCE modules, enforcing the framework to learn a more fine-grained, generalizable and pathologically discriminative medical visual representation. The overall training objectives is formulated as:
\begin{align}
\label{eq:overall_obj}
\mathcal{L} = \mathcal{L}_{ICMA} + \lambda \mathcal{L}_{PCMA} + \beta \mathcal{L}_{CCE},
\end{align} 
\noindent where $\lambda$ and $\beta$ are two hyper-parameters to balance the contribution between different objectives.

\section{Experiments}
\label{sec:experiments}
\subsection{Pre-training Setup}
\noindent \textbf{Dataset}: We follow most previous studies~\cite{cheng2023prior,huang2021gloria,wang2022multi} to pre-train our PLACE on the MIMIC-CXR dataset~\cite{johnson2019mimic} and adopt the same data pre-processing procedure as~\cite{wang2022multi,li2024mlip}. The images of the lateral view are removed since the downstream tasks only contain the frontal view images. Reports are formed by concatenating the Finding and Impression sections. We remove samples with an empty report or less than three words, resulting in a roughly total of 217,000 image-report pairs.  

\noindent \textbf{Implementation Details}:
Following~\cite{cheng2023prior,huang2021gloria,wang2022multi}, we employ ResNet-50~\cite{he2016deep} and ClinicalBERT~\cite{alsentzer2019publicly} as the backbone of our image and report encoder. Note that our proposed method is model-agnostic and can be applied to various backbones such as a vision transformer and convolutional networks. We also report the results of adopting the vision transformer as the image encoder in classification tasks. Images are first resized to $256\times256$ while maintaining the original size ratio with zero-padding for the smaller dimension, and then randomly cropped to $224\times 224$ during training. We adopt the AdamW~\cite{loshchilovdecoupled} as the optimizer with a learning rate of $4e-4$ and weight decay of $5e-2$. The batch size is set to 128 on three A100-40G GPU cards. We train our model for $50$ epochs with an early stop mechanism that terminates the training without seeing a decrease in validation loss for more than $10$ epochs. Consistent with~\cite{wang2022multi,chen2020simple}, we set the softmax temperature $\tau_1$ and $\tau_2$ to $0.07$ and $0.10$ respectively.
The split patch size $PS$ in CCP module is set to $32$ (determined by a small grid search in $\{16,28,32,56\}$, resulting in a $N_p$ of $49$. We set the number of pathological query tokens in~\autoref{eq:pca} to $12$. The loss weights $\lambda$ and $\beta$ are set to $0.5$ determined by a small grid search in $\{0.1,0.25,0.5,1\}$.

% \begin{table}
% 	%	\setlength{\belowdisplayskip}{1.5cm}
% 	\setlength{\abovedisplayskip}{-0.5cm}
% 	\centering
% 	\resizebox{\linewidth}{!}{
% 		\begin{tabular}{ccccccc}
% 			\toprule
% 			\multirow{2}*{Method} & \multicolumn{3}{c}{RSNA(Dice)} & \multicolumn{3}{c}{SIIM(Dice)}  \\
% 			&1\% & 10\% & 100\% & 1\% & 10\% & 100\%
% 			\\ 
% 			\midrule
% 			ConVIRT \cite{zhang2022contrastive}&55.0& 67.4& 67.5&25.0&43.2 &59.9\\
% 			GLoRIA-CheXpert \cite{huang2021gloria}&59.3& 67.5& 67.8&35.8 &46.9 & 63.4 \\
% 			GLoRIA-MIMIC \cite{huang2021gloria}&60.8& 68.2& 67.6&37.6& 56.4&64.0\\
% 			MGCA  \cite{wang2022multi}&{63.0}& {68.3}&{69.8} &{49.7} & {59.3}& {64.2}\\
%         M-FLAG  \cite{liu2023m}&{64.6}& {69.7}&{70.5} &{52.5} & {61.2}& {64.8}\\
%         PRIOR  \cite{cheng2023prior}&{66.4}& {68.3}&{72.7} &{51.2} & {59.7}& {66.3}\\
% 			MLIP  \cite{li2024mlip}&  {67.7}& {68.8}& {73.5}& {51.6} & {60.8}&{68.1}\\
%    PLACE (Ours)  &  {74.2}& {76.4}& {77.0}& {64.7} & {73.5}&{73.8}\\
% 			%			MLIP(Ours, ViT-B/16)
% 			\bottomrule
% 	\end{tabular}}
% 	\caption{Fine-tuned results of semantic segmentation under the setting of 1\%, 10\%, and 100\% of the training samples.}
% 	\label{tab:seg}

\subsection{Downstream Tasks}
Here, we outline the experimental setup for downstream tasks, {following the same downstream dataset and fine-tuning protocols described in previous works~\cite{wang2022multi,li2024mlip}. More detailed information can be found in these references.}

\noindent \textbf{Medical Semantic Segmentation.} The \textbf{SIIM} Pneumothorax~\cite{siim-acr-pneumothorax-segmentation} and \textbf{RSNA} Pneumonia~\cite{shih2019augmenting} datasets are used to assess the capability of our model for this task. Consistent with most previous works, we adopt the U-Net architecture and employ our pre-trained image encoder as the the encoder backbone (weight frozen), while fine-tuning the decoder using 1\%, 10\% and 100\% training samples. Dice score~\cite{wang2020image} is selected as the evaluation metric.

\noindent \textbf{Medical Object Detection.} The performance of our method for medical object detection is verified on \textbf{RSNA} Pneumonia (stage 2 version)~\cite{shih2019augmenting} and Object CXR~\cite{healthcare2020object} datasets. We utilise the YOLOv3 training protocol and set our pre-trained image encoder as a fixed backbone in the setting of 1\%, 10\% and 100\% training samples. The Mean Average Precision (mAP, IoU threshold from 0.4 to 0.75) is selected to gauge the model performance.

\noindent \textbf{Medical Image Classification.} We verify the effectiveness of our pre-trained image encoder for medical image classification on %\textbf{RSNA} Pneumonia~\cite{shih2019augmenting}, 
\textbf{COVIDx}~\cite{wang2020covid} and \textbf{CheXpert}~\cite{irvin2019chexpert} datasets. Following prior work~\cite{wang2022multi,li2024mlip,huang2021gloria}, we adopt linear probing which freezes the pre-trained image encoder and only trains the classification head on COVIDx dataset on three scenarios, i.e., 1\%, 10\% and 100\% training samples. Additionally, we also gauge the zero-shot generalization capability of our model on CheXpert and COVIDx datasets. Same as ~\cite{wang2022multi,li2024mlip}, we also report the results of taking the vanilla ViT~\cite{dosovitskiy2021an} as the image encoder.

\noindent \textbf{Zero-shot Medical Image-to-Text Retrieval.} We follow previous works~\cite{cheng2023prior,li2024mlip} to explore the performance of our method for zero-shot Medical Image-to-Text Retrieval on the CheXpert $5\times 200$ dataset~\cite{cheng2023prior}. This task examines whether the model can retrieve reports that corresponded with the disease label of the query image. The performance of the model is assessed through the Precision@K measure. 

\noindent {\textbf{Medical Report Generation.} We further investigate PLACE's ability to comprehend cross-modal information on the MIMIC-CXR dataset through the task of report generation. We adopt two fundamental and typical architecture as our baseline (1) ST~\cite{vaswani2017attention}: a visual extractor with transformer-based encoder-decoder architecture and (2) C2GPT2~\cite{nicolson2023improving}: a vision encoder-language decoder architecture. Note that the baselines are standard transformer-based models without any advanced techniques or additional data sources tailored specifically to MRG. In our approach, we employ the pretrained image encoder from PLACE as the visual extractor in ST and as the vision encoder in Res2GPT2, maintaining fixed weights while only optimizing other model components. The official data split is adopted. We evaluate the performance of the model by the commonly utilized metrics including BLEU~\cite{papineni2002bleu}, METEOR~\cite{denkowski2011meteor}, ROUGE-L~\cite{lin2004rouge} and CIDEr~\cite{vedantam2015cider}.
}

\begin{table}
	\setlength{\belowdisplayskip}{-5cm}
	\centering
	\resizebox{\linewidth}{!}{
		\begin{tabular}{ccccccc}
			\toprule
			\multirow{2}*{Method} & \multicolumn{3}{c}{RSNA(mAP)} & \multicolumn{3}{c}{Object CXR(mAP)}  \\
			& 1\% & 10\% & 100\% &1\% & 10\% & 100\%
			\\ 
			\midrule
			%Random Init & 1.0&4.0& 	8.9&$\sim$&$\sim$& 4.4 \\
			%ImageNet Init & 3.6 &8.0 & 15.7 &$\sim$& 8.6& 15.9 \\
			ConVIRT \cite{zhang2022contrastive} &8.2& 15.6& 17.9& $\sim$&8.6&15.9\\
			GLoRIA-CheXpert \cite{huang2021gloria}&9.8& 14.8& 18.8&$\sim$&10.6 &15.6 \\
			GLoRIA-MIMIC \cite{huang2021gloria}& 10.3& 15.6&23.1&$\sim$& 8.9&16.6\\
			MGCA \cite{wang2022multi}& {12.9}& {16.8}&{24.9}  & $\sim$&{12.1}& {19.2}\\
            M-FLAG \cite{liu2023m}& {13.7}& {17.5}&{25.4}  & $\sim$&{13.6}& {19.5}\\    
            PRIOR \cite{cheng2023prior}& {15.6}& {18.5}&{25.2}  & 2.9&15.2& {19.8}\\ 
		MLIP~\cite{li2024mlip}& {17.2}& {19.1}& {25.8}& {4.6}&\textbf{17.4} & {20.2}\\
        G2D~\cite{liu2024g2d} & {15.9}& {21.7}&{27.2}  & 3.8&13.1& {20.4}\\ 
   		 PLACE (Ours) & \textbf{22.4}& \textbf{21.8}& \textbf{28.7}& \textbf{10.0}&{16.1} & \textbf{20.6}\\
			\bottomrule
	\end{tabular}}
	\caption{Fine-tuned results of object detection under the setting of 1\%, 10\%, and 100\%. $\sim$ means mAP is smaller than 1\%.}
	\label{tab:det}
\end{table}

\begin{table}
  \centering
  \vspace{2mm}
  \resizebox{0.4\textwidth}{!}{%
  \begin{tabular}{lcccc}
    \toprule
  \multirow{2}{*}{VLP Methods} & \multicolumn{4}{c}{CheXpert Image-to-text Retrieval} \\ \cline{2-5} 
    & Prec $@$ 1   & Prec $@$ 2  & Prec $@$ 5  & Prec $@$ 10   \\ \hline
  ConVIRT \cite{zhang2022contrastive}   &  20.3   &   19.8 &  19.7&19.9               \\
 GLoRIA  \cite{huang2021gloria} &{29.3}&{29.0}&{27.8}& {26.8}               \\
 {PRIOR} \cite{cheng2023prior}   &{{40.2}}&{{39.6}} & 39.3 &{ {38.0}}   
 \\ 

  {MLIP} \cite{li2024mlip}  & 41.7 &  40.3   &  39.0     &   {39.4}   
 \\ 

  {MGCA} \cite{wang2022multi}  & {42.5} &   {41.9}                &   {40.5}       &   {39.4}   
 \\ 
   PLACE (Ours)  & \textbf{44.8} &   \textbf{44.8}  &  \textbf{43.8}  &   \textbf{42.5}   
 \\ 
 
 \bottomrule
  \end{tabular}%
  }
\caption{Zero-shot Image-to-text retrieval results.}
  \label{tab:retrieval}
  \end{table}

  \begin{table*}
	\setlength{\belowdisplayskip}{0.0cm}
	\centering
	\resizebox{\linewidth}{!}{
		\begin{tabular}{ccc|cccccc|cccccc|ccc}
			\toprule
			\multicolumn{3}{c}{Learning Objective} & \multicolumn{3}{|c}{RSNA(Dice)} & \multicolumn{3}{c|}{SIIM(Dice)} & \multicolumn{3}{c}{RSNA(mAP)} & \multicolumn{3}{c|}{Object CXR(mAP)} & \multicolumn{2}{c}{COVIDx(ACC)} \\
			ICMA& CCE &PCMA &1\% & 10\% & 100\% & 1\% & 10\% & 100\% &1\% & 10\% & 100\% & 1\% & 10\% & 100\% &0-shot  &100\%
			\\ 
			\hline
			\checkmark & & &69.6&73.2&73.3&54.6&68.4&69.8 &14.9 &20.2 &26.1&2.4 &13.1 &18.8 &36.5 &92.8\\
			\checkmark&\checkmark&&71.8&74.2&74.1&59.1&71.5&72.9 &17.5 &21.5 &27.8&3.8 &15.6 &19.6 &38.8 &93.5 \\
			\checkmark&&\checkmark&73.3&75.8&75.0&60.5&72.7&72.5 &21.0 &20.6 &27.3&5.6 &15.4 &20.4 &40.3 &93.5\\
    \checkmark&\checkmark&\checkmark&\textbf{74.2}&\textbf{76.4}&\textbf{77.0}&\textbf{64.7}&\textbf{73.5}&\textbf{73.8} &\textbf{22.4} &\textbf{21.8} &\textbf{28.7}&\textbf{10.0} &\textbf{16.1} &\textbf{20.6} &\textbf{44.0} &\textbf{94.0}\\
			\bottomrule
	\end{tabular}}
	\caption{Ablation study of our model on semantic segmentation, object detection and image classification tasks.}
	\label{tab:ablation_study}
\vspace{-10pt}
\end{table*}

\subsection{Results}
\label{sec:results}

\noindent \textbf{Results on Classification and Image-to-text Retrieval.} Results in \autoref{tab:seg_and_cls} demonstrate the effective of PLACE on image classification tasks where our model achieves the best performance on both the fine-tuned settings and zero-shot scenario, outperforming the second-best results by a significant margin. Moreover, PLACE improves remarkably the zero-shot classification performance on both the CheXpert and COVIDx dataset. A similar pattern is shown in the image-to-text retrieval task in \autoref{tab:retrieval} where our model obtains the highest scores for all the setting of $K$, indicating  better capability of aligning the pathology information. These results, especially for the zero-shot scenarios, further confirm the efficacy of PLACE.

\begin{table}

	\setlength{\belowdisplayskip}{-5cm}
	\centering
	\resizebox{0.95\linewidth}{!}{
\label{tab:rrg}
\begin{tabular}{l|l|ccccc}
\toprule  
\textbf{Group} &\textbf{Method}  & \textbf{BL1}  & \textbf{BL4} &\textbf{MTOR} &\textbf{RG-L} &\textbf{CDr}  \\
\midrule  
\multirowcell{7}{Specific \\to MRG} &CACRG~\cite{liu2019clinically} &0.313  &0.103 &- &0.306 &-  \\
& XPRONet~\cite{wang2022cross} & 0.344   & 0.105 & 0.138  &0.279 &0.154 \\
& DCL~\cite{li2023dynamic} &-  & 0.109  & 0.150 & 0.284  &0.281   \\
& UAR~\cite{li2023unify} &0.363 & 0.107  &0.157  & 0.286  &0.246   \\
&MCSAM~\cite{tao2024memory} &0.379  & 0.109  &0.149& 0.284 &-   \\
&KCAP~\cite{huang2024knowledge} &0.378   & 0.121  &0.149& 0.301 &-   \\
&ATL-CA$_C$~\cite{mei2025adaptive} &0.382  & 0.138  &0.157  & 0.321 &0.239  \\
\hline
\multirowcell{7}{General \\VLP} &Med-Flamingo~\cite{moor2023med} &0.233   &0.019 &0.080 &0.123  &-  \\
%& Llava-med~\cite{li2023llava} &0.199 &0.006 &0.163  &0.097 &  &- &- &-   \\
& Uni-Med~\cite{zhuuni}  &0.278 &0.065 &0.106 &0.226 &-    \\
& PTUnifier~\cite{chen2023towards}  &- &0.107 &- &0.210 &-   \\
& BioViL-T~\cite{bannur2023learning}  &-  &0.092 &- &0.296 & -  \\
\cline{2-7}
& ST (ImageNet)  &0.334  &0.098 &0.128&0.267 &0.200  \\
& ST (Our-PLACE)  &\textbf{0.361}  &\textbf{0.109} &\textbf{0.140} &\textbf{0.276} &\textbf{0.270} \\
& C2DG2 (ImageNet)  &0.360  &0.105 &0.134 &0.266& 0.256 \\
& C2DG2 (Our-PLACE)  &\textbf{0.387}  &\textbf{0.118} &\textbf{0.149} &\textbf{0.278} &\textbf{0.357}  \\

\bottomrule 

\end{tabular}}
\caption{Results of report generation in the MIMIC-CXR dataset. BL, MTOR, RG-L, CDr are the abbreviations of BLEU, METEOR, Rouge-L and CIDEr respectively. }
\vspace{-10pt}
\end{table}

\noindent \textbf{Results on Semantic Segmentation.}
We report the results of semantic segmentation in \autoref{tab:seg_and_cls}. As can be seen, PLACE surpasses previous methods by a notable margin on both the RSNA and SIIM datasets. Notably, the superiority of PLACE becomes more pronounced in scenarios characterized by limited data availability, e.g., 1\% and 10\% of the training samples, suggesting the efficacy of our approach in learning highly generalizable, fine-grained representations, which is particularly crucial in small-data regimes on tasks with higher demand for localized and fine-grained features.

\noindent \textbf{Results on Object Detection.}
\autoref{tab:det} shows the results of object detection on the RSNA and Object-CXR datasets. Our method, PLACE, demonstrates a significant improvement over the previous methods under all the settings except for the 10\% training sample on Object-CXR dataset where our model obtains a slightly lower score than MLIP. Notably, PLACE shows more obvious superiority over the previous methods on the small data regime, and our model trained with 1\% samples even achieves slightly higher score than 10\% setting on the RSNA dataset.

\noindent {\textbf{Results on Report Generation.} We compare the efficacy of PLACE against recent RRG-specific models (including VLP specific to MRG), and general Visual Language Pretraining (VLP) approaches. Furthermore, to underscore the efficacy of PLACE, we present results of the same models with visual extractor/encoder weights initialized from ImageNet pre-trained models. As illustrated in~\autoref{tab:rrg}, our approach significantly outperforms previous VLP methods and demonstrates superior performance compared to models initialized with ImageNet weights. These findings highlight the cross-modal capabilities of PLACE and its effectiveness in learning a fine-grained and pathology-enriched visual representation. Moreover, even when utilizing a frozen weight setting, our model achieves competitive results compared to most models tailored to MRG tasks. It is noteworthy that our evaluation of the effectiveness of PLACE is based on standard transformer-based baselines, without incorporating advanced MRG-specific techniques or additional knowledge sources. Despite these achievements, there remains a performance disparity between general VLP models and those designed specifically for MRG tasks. This could be attributed to two main factors. Firstly, the data volume during the pre-training phase is comparable to that of the MRG task since both utilize the MIMIC-CXR dataset, leading to limited additional knowledge transfer when adapting general VLP models to MRG tasks. Secondly, MRG-specific methods often leverage supplementary information or data sources such as disease labels and knowledge graphs. For example, ATL-CA$_C$ incorporates crucial clinical history, such as comparison and indication, as auxiliary inputs to enhance model performance. %In the future, we aim to further investigate the performance of general VLP models in generative tasks like MRG.
}

\subsection{Ablation Studies}
\label{sec:ablation}

\subsubsection{Contribution of each component}
We investigate the contribution of each proposed module, including the Pathological-level Cross-modal Alignment (PCMA) and Cross-modal Correlation Exploration (CCE) on semantic segmentation, object detection, and image classification tasks. The model optimized only by instance-level cross-modal alignment (ICMA) is considered as the baseline. As shown in \autoref{tab:ablation_study}, each stage of our design shows consistent improvements over the baseline model. Additionally, further combining the PCMA and CCE modules brings more notable gains for downstream tasks, especially for challenging ones such as semantic segmentation requiring fine-grained details and zero-shot classification. It is noteworthy that our method shows higher efficacy in the setting when fewer training samples available. For example, our full PLACE model enhances the performance of baseline by 4.6 (RSNA) and 10.1 (SIIM) on the segmentation task in the 1\% training scenario, and significantly improves the accuracy of the zero-shot classification from 36.5\% to 44.0\%. These results show that PLACE can learn a more generalizable and fine-grained visual representation, confirming the effectiveness of our proposed methods. We attribute the improvements mainly to the better exploitation of the naturally exhibited pathological correspondences across image and reports, and to the exploration of the fine-grained details.

\subsubsection{Additional exploration of the PCMA objective}
In addition to the ablation study presented in the previous section, we conduct a more comprehensive investigation of the effectiveness of our proposed PCMA objective by comparing our complete PLACE with two variants: (1) PLACE without V-PORs: in the PCMA module, we remove the \textit{V-PORs}, thus downgrading the alignment to the level between the original localized visual tokens and T-PORs; (2) PLACE without T-PORs: in the PCMA module, we remove the \textit{T-PORs}, thus downgrading the alignment to the level between the original localized visual tokens and T-PORs. As evidenced in \autoref{tab:ablation_variants}, the exclusion of V-PORs or T-PORs results in diminished performance, and a more significant decline can be observed in more complex dense prediction tasks, such as semantic segmentation. These findings corroborate the efficacy of our V-PORs design and the proposed alignment module (namely PCMA Align) on the pathological level. Moreover, they underscore the significance of augmenting the granularity from image patches $\leftrightarrow$ words to pathological regions $\leftrightarrow$ pathological sentences (the textual pathological observations).
% from image patches $\leftrightarrow$ words to a pathological level.

\begin{table}
	\setlength{\belowdisplayskip}{-5cm}
	\centering
	\resizebox{0.95\linewidth}{!}{
		\begin{tabular}{ccccccc}
			\toprule
			\multirow{2}*{Method} & \multicolumn{3}{c}{RSNA(Dice)} & \multicolumn{3}{c}{COVIDx(ACC)}  \\
			& 1\% & 10\% & 100\% &1\% & 10\% & 100\%
			\\ 
			\midrule
			%Random Init & 1.0&4.0& 	8.9&$\sim$&$\sim$& 4.4 \\
			%ImageNet Init & 3.6 &8.0 & 15.7 &$\sim$& 8.6& 15.9 \\

        PLACE $w/o$ V-PORs &71.8& 73.6& 73.9& 76.5&88.3&93.5\\

     PLACE $w/o$ T-PORs & 69.4& 74.3&75.4&76.0& 88.3&93.8\\

        \textbf{PLACE} &\textbf{74.2}& \textbf{76.4}& \textbf{77.0}&\textbf{76.8}&\textbf{89.3} &\textbf{94.0} \\
    \bottomrule
	\end{tabular}}
	\caption{Comparison of our full PLACE with two variants of PLACE.}
	\label{tab:ablation_variants}
\end{table}

\subsubsection{Influence of the number of pathology query tokens.} We investigate the influence of PLACE to the number of pathology query tokens $N_p$ by varying $N_p$ from $10$ to $16$ on semantic segmentation and classification tasks. The results in \autoref{tab:ablation_np} show that PLACE is relatively robust to $N_p$. Nevertheless, it is still beneficial to set an appropriate value to achieve the best performance, since $N_p$ represents the number of pathological observations that the model will extract from the images. A too large a value exceeding the total pathological observations may introduce noisy information, while a too small a value struggles to cover all the pathological observations.

\begin{table}
	\setlength{\belowdisplayskip}{-5cm}
	\centering
	\resizebox{0.8\linewidth}{!}{
		\begin{tabular}{ccccccc}
			\toprule
			\multirow{2}*{$N_q$} & \multicolumn{3}{c}{RSNA(Dice)} & \multicolumn{3}{c}{COVIDx(ACC)}  \\
			& 1\% & 10\% & 100\% &1\% & 10\% & 100\%
			\\ 
			\midrule
			%Random Init & 1.0&4.0& 	8.9&$\sim$&$\sim$& 4.4 \\
			%ImageNet Init & 3.6 &8.0 & 15.7 &$\sim$& 8.6& 15.9 \\
			10 &73.8& 74.2& 73.9& \textbf{77.0}&87.5&93.3\\
			\textbf{12} &\textbf{74.2}& \textbf{76.4}& \textbf{77.0}&76.8&\textbf{89.3} &\textbf{94.0} \\
			 14 & 72.6& 73.6&74.8&76.8& 88.3&91.8\\
			16 & {70.9}& {75.0}&{73.0}  & 75.5&{88.0}& {93.8}\\
			\bottomrule
	\end{tabular}}
	\caption{Effect of varying $N_q$, the number of pathology query tokens on semantic segmentation and classification tasks.}
	\label{tab:ablation_np}
    \vspace{-10pt}
\end{table}

\subsubsection{Influence of patch size in Correlation Exploration.} We explore the influence on PLACE of the patch size $PS$ in the CCE module, on semantic segmentation and classification tasks. Note that the image size needs to be wholly divisible by $PS$ to ensure that the total number of patches is an integer. We conduct the experiments on a $PS$ of $16,28,32,56$, resulting in a total of $196$,$64$,$49$ and $16$ image patches attending the calculation of covariance matrix. As can be seen in~\autoref{tab:ablation_ps}, PLACE is not overly sensitive to the value of the patch size. Nonetheless, a smaller value may disrupt the integrity of semantic information, potentially compromising the efficacy of the covariance matrix in serving as a correlation descriptor. Conversely, excessively large values may introduce non-discriminative noise. Hence, an appropriate value of $PS$ can bring more significant performance to PLACE. 

\begin{table}
	\setlength{\belowdisplayskip}{-5cm}
	\centering
	\resizebox{0.8\linewidth}{!}{
		\begin{tabular}{ccccccc}
			\toprule
			\multirow{2}*{$PS$} & \multicolumn{3}{c}{RSNA(Dice)} & \multicolumn{3}{c}{COVIDx(ACC)}  \\
			& 1\% & 10\% & 100\% &1\% & 10\% & 100\%
			\\ 
			\midrule
			%Random Init & 1.0&4.0& 	8.9&$\sim$&$\sim$& 4.4 \\
			%ImageNet Init & 3.6 &8.0 & 15.7 &$\sim$& 8.6& 15.9 \\

        16 &71.9& 74.3& 77.2& 76.5&86.3&92.3\\

     28 & 73.4& 70.7&77.5&\textbf{77.5}& 87.3&91.8\\

        \textbf{32} &\textbf{74.2}& \textbf{76.4}& \textbf{77.0}&76.8&\textbf{89.3} &\textbf{94.0} \\

    56 & \textbf{74.2}& {75.4}&{73.0}  & 76.8&{88.0}& {93.8}\\
    \bottomrule
	\end{tabular}}
	\caption{Effect of varying $PS$, the patch size in the Cross-Modal Correlation Exploration Module.}
	\label{tab:ablation_ps}
\end{table}

\subsection{{Computational Complexity Analysis}}
 
{Here, we analyze the parameter count of the \textit{Base} model and the proposed \textit{PLACE} models as shown in \autoref{tab:mp}. The base model comprises solely the image and text encoders. By incorporating the full PLACE module (Base+PCMA+CCE), an increase of 13.0M learnable parameters is observed. Despite the rise in computational complexity, it remains relatively insignificant compared to the base model (113.4M), while significantly enhancing model performance across various downstream tasks. Additionally, the integration of the CCE module introduces only 0.9M learnable parameters to the models, which is negligible in comparison to the base model, yet leads to notable enhancements in tasks such as medical detection and segmentation. }

{The CCE module is constructed with a linear layer sized $D \times K$, where $D$ represents the dimension of the global report representation (768 here) and $K$ signifies the number of elements in the covariance matrix. It is noteworthy that the covariance matrix assesses the correlation between each unique pair of patches. As a result, the value of $K$ increases quadratically with the number of image patches, which in turn is also quadratic to the patch size with a fixed image resolution. For example, given the image resolution of 224 and the patch size $P$, the total number of element in covariance matrix $K=(\lfloor\frac{224}{P}\rfloor)^4$. Consequently, the number of parameters in the CCE module grows inversely proportional to the fourth power of the patch size. Given that the covariance matrix is symmetric, it is only necessary for the model to predict the upper triangular portion of the matrix (correlation for image patch to itself is also excluded). This reduces the parameter count by half to $\{\frac{K}{2}-{num\_patches}\}$ in our implementation. Considering a small patch size of 16, this results in a total of 14.7M learnable parameters. However, as indicated in the ablation study in~\autoref{sec:ablation} (4), the model is not overly sensitive to variations in patch size. It is generally advised against choosing very small patch sizes like 8 or 16 due to the risk of disrupting semantic continuity. Conversely, selecting a very large patch size, for instance, 112, may compromise the exploration of fine-grained local information. Therefore, it remains crucial to choose an appropriate patch size to obtain the optimal performance and balance better performance with higher efficiency. A commonly adopted configuration in various downstream applications, i.e., the one used in our study, is with a patch size of $32$ and an image resolution of $224 \times 224$, introduces only 0.9M parameters.}

%{%For example, in the setting of this paper, we set the patch size to $32$ with an input image of $224\times 224$ resolution, leading to $7\times 7=49$ image patches and $49\times 49=2401$ elements in the covariance matrix. By predicting only the upper triangular portion and removing the self-correlation, $K=(49\times49-49)/2 = 1176$, hence the total number of parameter for the CCE module in our work is $D\times K = 768 \times 1176 = 903,168 (0.9M)$. Considering a small patch size of 16, this results in 
%196 image patches ($14 \times 14 = 196$), 38,220 elements in the covariance matrix ($196 \times 196 = 38,220$), and
%a total of 14.7M learnable parameters in the CCE module. }

\begin{table}
\begin{center}
\begin{tabular}{|c|c|c|c|}
\hline
 Method &Base & Base+PCMA  & Base+PCMA+CCE\\
\hline
\#Param (M) &113.4 & 125.5 (+12.1) & 126.4 (+13.0) \\
\hline
\end{tabular}
\end{center}
\caption{The number of parameters of different models.}
\label{tab:mp}
\vspace{-10pt}
\end{table}

\section{Further Analysis and Visualizations}
In this section, we provide additional analyses and visualizations to substantiate the efficacy of our methodologies. These visualisations also indicate PLACE to have good interpretability.
\subsubsection{Visual pathology observations}
\begin{figure}
\centering
\includegraphics[width=.99\linewidth]{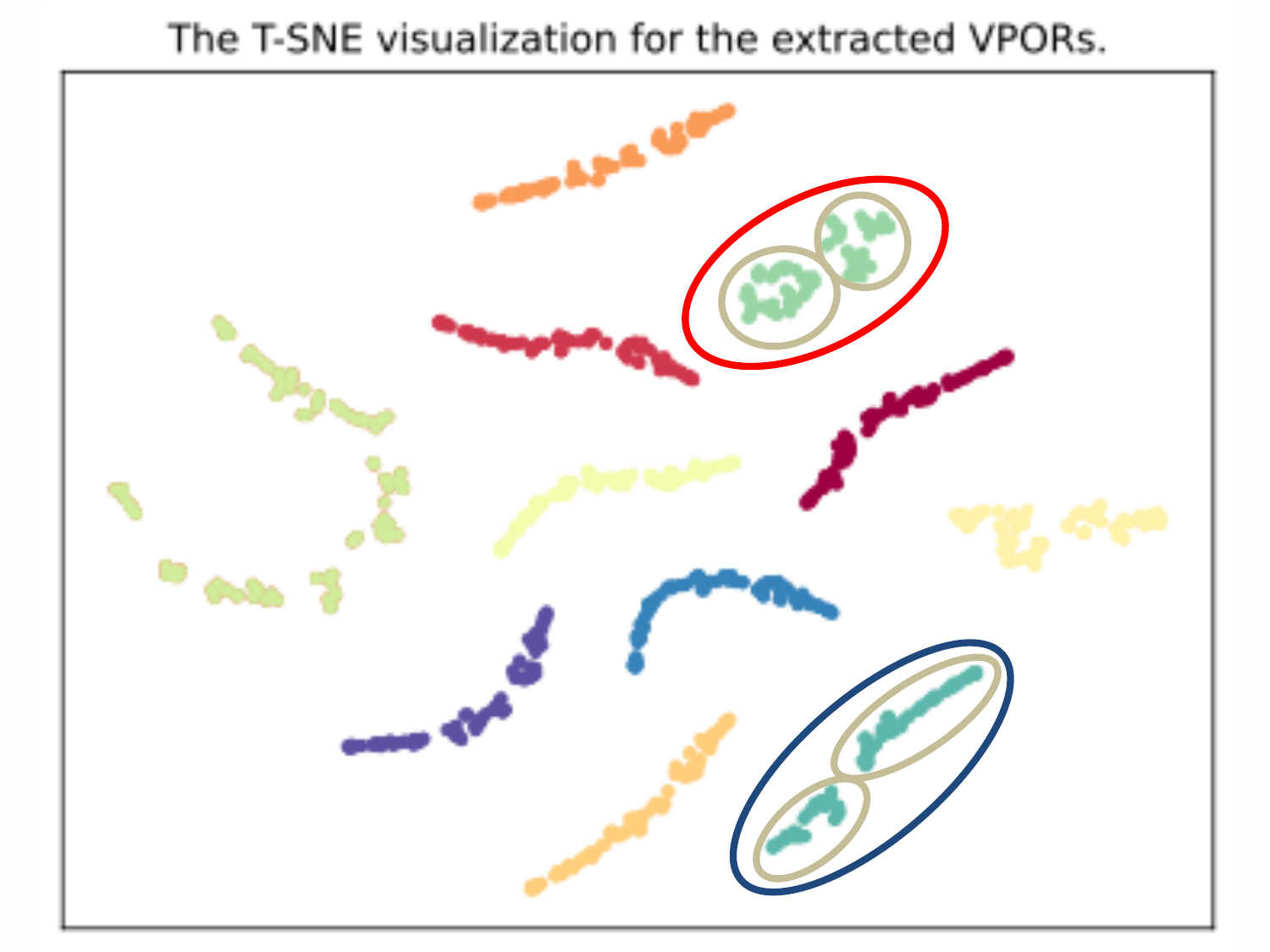}
\caption{{T-SNE visualisation for the extracted VPORs from 100 randomly selected samples in the test set.}}
\label{fig:tsne}
\end{figure}

\begin{figure}
\centering
\includegraphics[width=.9\linewidth]{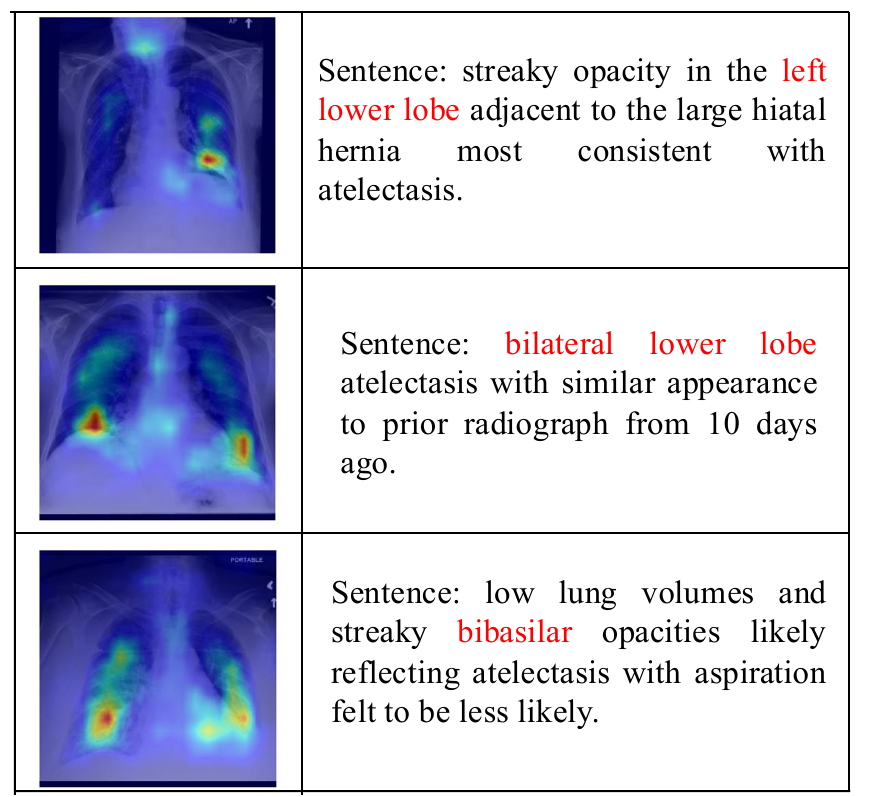}
\caption{{Visualization of the attention map of the V-PORs to the local visual tokens. These V-PORs are those having the highest similarity to the associated atelectasis-related T-PORs. The highlighted visual tokens are regarded as important regions learnt by the model.}}
\label{fig:vpors}
\end{figure}

To further explore whether PLACE can truly extract useful visual pathology observations, we present a T-SNE visualisation of visual pathology observation representations from 100 samples in~\autoref{fig:tsne} where points of the same colour are retrieved from the same pathology query token. Obviously, V-PORs retrieved by the same pathology query token are clustered, indicating that each is specifically functioned to extract one type of pathology from the visual tokens as expected. It is worth noting that in addition to the large clusters for different pathogenesis, each pathological-level cluster contains several sub-clusters, aligning to real-world situations that each pathology may contain multiple patterns such as normal, abnormal, and those of different severities. {To substantiate that these V-PORs can effectively extract information related to atelectasis from the images, we present visualizations of the attention scores between the V-POR that exhibits the highest similarity to the atelectasis-related T-PORs, and every local visual token produced by the cross-attention module in the VPOE shown in~\autoref{fig:vpors}. Evidently, these V-PORs successfully activate the regions associated with atelectasis in the image. Furthermore, they even identify precise locations as referenced in \textit{left lower lobe, bilateral lower lobe and bibasilar}. The preceding results unequivocally demonstrate that each pathology query token possesses the ability to extract pathology observations from the images and exhibit pathological-level alignment through the meticulous design of our VPOE and PCMA modules, thereby further confirming our theory.}

\begin{figure}
\centering
\includegraphics[width=.99\linewidth]{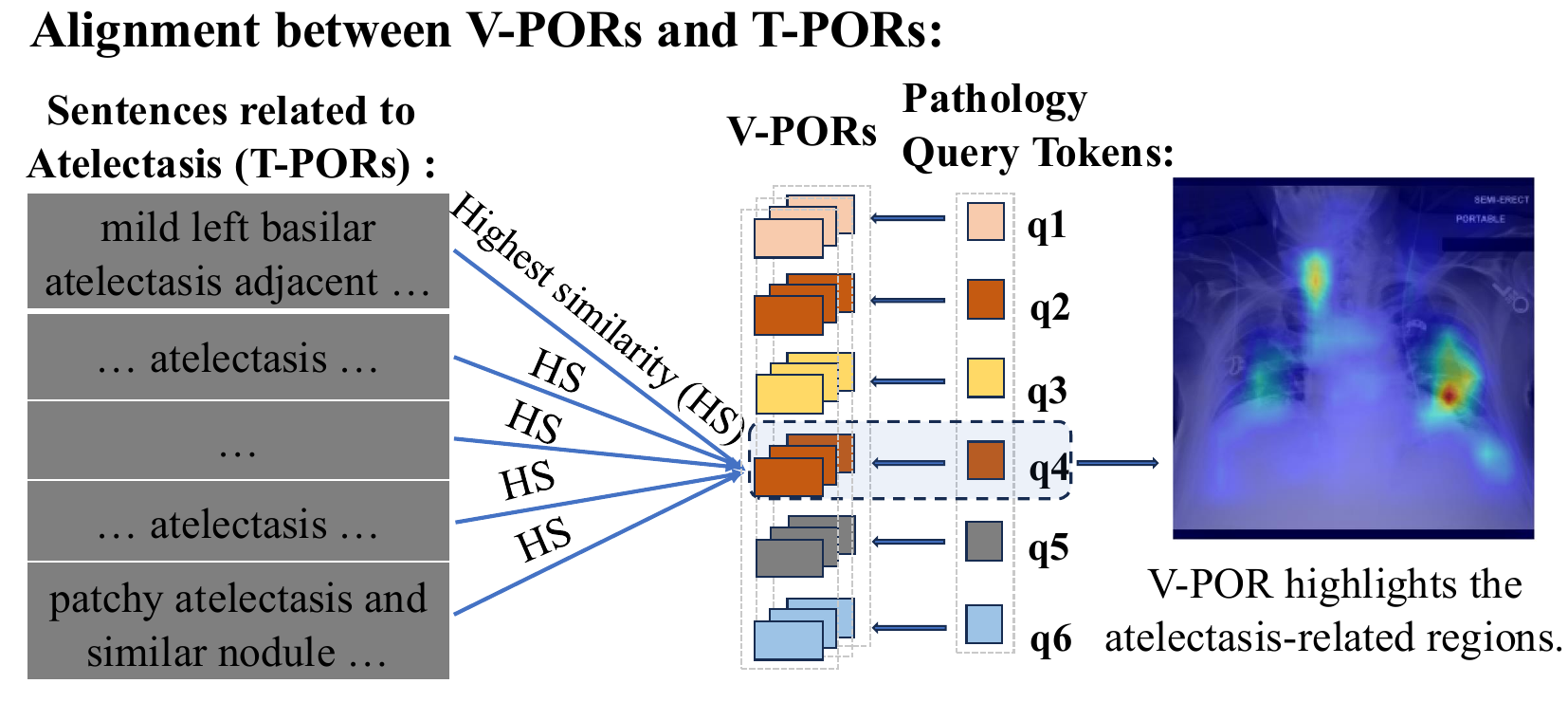}
\caption{{{An illustration of the pathological-level cross-modal alignment between the V-POR and T-POR. Each sentence (row) refers to a pathology observation from one sample. The proposed Visual Pathology Observation Extractor extracts a group of V-PORs for each image through the same learnable pathology query tokens.}}}
\label{fig:illustration}
\vspace{-10pt}
\end{figure}

\subsubsection{V-PORs and Pathological-Level Alignment}
\label{sec:pqt}
In addition to the T-SNE visualization of Visual Pathology Observation Representations (V-PORs), here, we further explore whether each V-POR can truly extract one or multiple observations from the images based on our proposed PCMA objective. {We break down the validation process into distinct steps, as depicted in Figure~\ref{fig:illustration}.}

{
\begin{enumerate}
    \item[A.] The query tokens facilitate the extraction of consistent V-PORs across various images $\&$ The V-PORs concentrate on the same anatomical regions $\rightarrow$ The V-PORs demonstrate an ability to capture pathological observations pertinent to a specific anatomical location. 
    \item[B.] The V-PORs extracted by the same query token exhibit significant similarity to the same type of T-PORs among the various T-PORs within each report. 
    \item[C.] The query tokens effectively extract V-PORs that are most relevant to the T-PORs. Given that the T-PORs, learned directly from the text, are clinically significant, one may conclude that the proposed PCMA module functions effectively, i.e., A$\&$B $\rightarrow$ C.
\end{enumerate}
}

%As demonstrated in~\autoref{fig:illustration}, when pathological-level alignment functions effectively, sentences which pertain to similar pathological observations are expected to exhibit maximal similarity with the same type of visual pathological observation representation (V-POR) extracted by the same pathology query token. This V-POR, extracted by a particular pathology query token, will exhibit sensitivity to the relevant regions within the image. 
The analysis in the previous paragraphs has confirmed the condition A. To validate B, we randomly select 200 samples showing the presence of the~\textit{Atelectasis} and locate the relevant sentences describing the~\textit{Atelectasis} observation by checking whether the sentence contains the pathology observation name\footnote{Samples without comprising the observation name are simply removed.}. Subsequently, we record the index of the V-POR that demonstrates the greatest similarity with each selected T-POR from the entire set of samples. Following this, we evaluate the consistency of the selected V-POR across all samples by calculating the proportion of the most frequently selected V-POR (top-1 result). The findings indicate that $47.4\%$ of the samples exhibit a consistent top-one-ranked V-POR. Acknowledging that certain sentences may encompass multiple observations, e.g., sentence ``\textit{extremely low lung volumes with pulmonary vascular crowding and left basilar atelectasis}," we additionally present the top-2 result ($59.3\%$), which includes samples wherein the most frequently selected V-POR ranks second. The preceding results unequivocally demonstrate that the query tokens possess the ability to extract pathology observations from the images and exhibit pathological-level alignment with our PCMA module, thereby further substantiating the validity of our framework.

\subsubsection{Attention Map for Images}
\label{sec:attn_map_img}
\begin{figure}
\centering
\includegraphics[width=.8\linewidth]{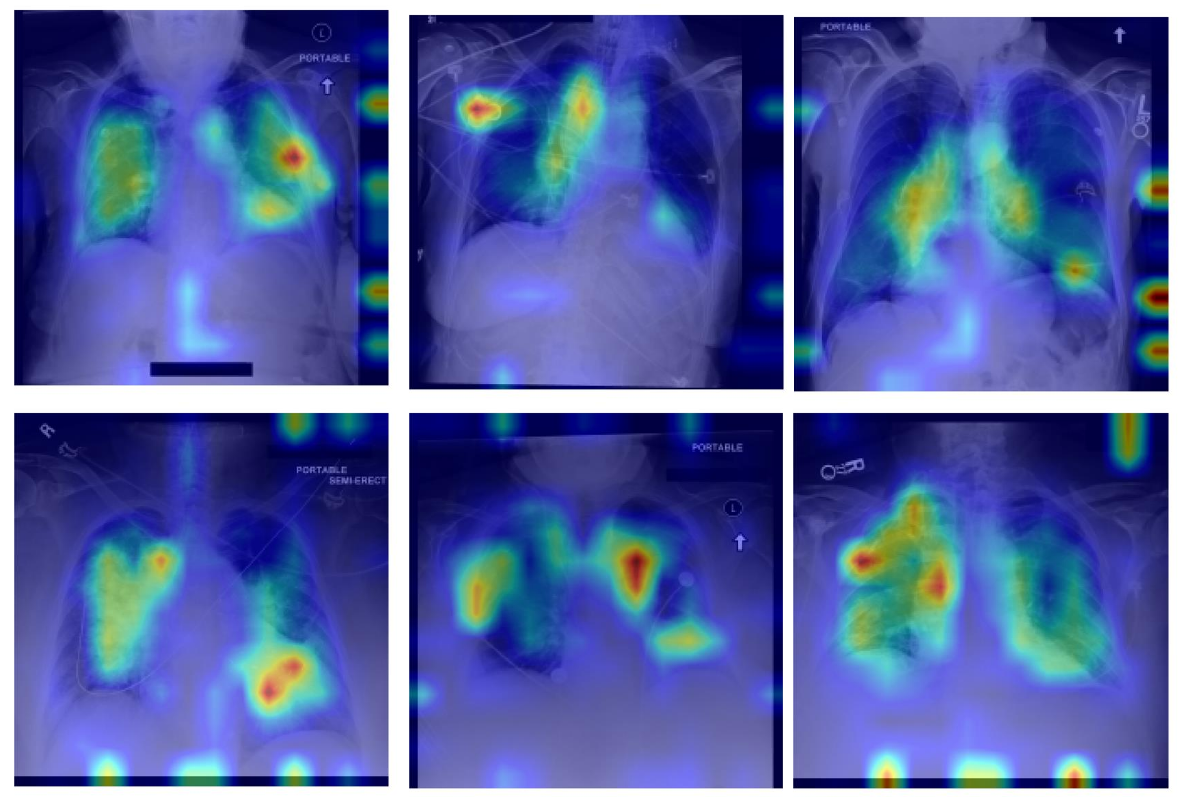}
\caption{Visualization of the attention map for visual tokens. The highlighted areas are regarded as important regions learnt by the model.}
\label{fig:img_attn}
\vspace{-10pt}
\end{figure}

We provide visualizations of the attention maps from the ViT-based PLACE in~\autoref{fig:img_attn} to understand the importance of each visual token when learning the global image representation. In particular, we randomly select six samples and visualize the attention scores of the $[CLS]$ token to other visual tokens from the last layer of ViT. These activated regions are identified as important to the task by the model through the training. This visualization result shows that our model is capable of localizing regions, e.g., lungs and hearts, that are crucial to the understand the task.

\subsubsection{Visualization of Important Words}
\label{sec:vis_words} We present the ten most significant words (highlighted in bold) from four sample reports learned by the model in \autoref{tab:imp_words}. These words are identified by averaging the attention weights from BERT's final layer. %across multiple heads
As can be seen, the majority of these highlighted words, i.e., \textit{consolidation, pneumonia, pleural, minimal}, pertain to patients' medical conditions. Furthermore, our model demonstrates a capability of discerning the location and severity of pathological observations, as illustrated by words such as \textit{mild, stable, lower, left, right}. These findings suggest that our proposed framework, PLACE, is capable of acquiring a more fine-grained and pathologically enriched representation.

\begin{table*}
	\setlength{\belowdisplayskip}{-5cm}
	\centering
		\begin{tabular}{p{16cm}}
        \toprule

\textbf{Report:} worsened bilateral lower lung \textbf{consolidation} concerning \textbf{for} \textbf{pneumonia} in the appropriate clinical setting. \textbf{the} right and left \textbf{lower} lung have increased consolidation since prior examination. \textbf{no} pneumothorax. cardiomediastinal is otherwise unchanged as compared to previous examination. cardiac leads are seen \textbf{terminating} in the presumed \textbf{right} atrium and \textbf{right} ventricle. \\
\hline
        
\textbf{Report:} improved mild pulmonary edema. \textbf{the} endotracheal \textbf{tube} has been \textbf{removed}. a single lead {external} pacer remains in place. there is \textbf{no} pneumothorax. \textbf{mild} cardiomegaly despite the projection is unchanged. the patient has had previous aortic valve replacement. mild pulmonary edema has improved . \textbf{minimal} \textbf{retrocardia} \textbf{subsegmenta} atelectasis \textbf{is} unchanged. \\
        \hline

   \textbf{Report:} no \textbf{acute} cardiopulmonary process. \textbf{stable} \textbf{bibasilar} atelectasis or scarring. a single frontal view of the chest shows \textbf{linear} opacities at the bilateral bases greater on the right than the left. these are stable from prior exams and most consistent \textbf{with} atelectasis or scarring. no new opacities identified. the lung volumes are low. there is no \textbf{pulmonary} \textbf{edema} \textbf{pleural} effusion \textbf{or} pneumothorax. \textbf{the} cardiomediastinal silhouette is normal.  \\
   \hline
	\textbf{Report:} 
    % \textbf{new} heterogeneous consolidation in the left lower lobe is concerning for pneumonia in the appropriate clinical setting. the cardiac silhouette is unremarkable . again noted is \textbf{bilateral} \textbf{hilar} lymphadenopathy. two \textbf{central} \textbf{venous} catheters remain in stable position. there is \textbf{new} heterogeneous consolidation in the left lower \textbf{lobe} accompanied by a small {left} pleural effusion. pulmonary vascular congestion is accompanied by \textbf{minimal} \textbf{interstitial} edema. 
    persistent \textbf{right} pleural effusion. persistent right lower lobe opacity. clinical correlation for signs of continued or recurrent infection is recommended ct could be performed for further evaluation as clinically indicated. findings and recommendations were reported to \textbf{the} radiology \textbf{communication} dashboard on X. small to moderate \textbf{right} \textbf{pleural} effusion \textbf{has} minimally decreased compared to prior. there is somewhat \textbf{improved} aeration at the right \textbf{lung} base with persistent right lower lobe opacity. no new consolidation \textbf{left} \textbf{pleural} effusion.
    \\
        \bottomrule
	\end{tabular}
	\caption{An illustration of the top ten important words of four samples learnt by the text encoder of PLACE during the training process. The important words are highlighted in bold.}
	\label{tab:imp_words}
    \vspace{-10pt}
\end{table*}
\section{Conclusions}
\label{sec:concluison}
This paper introduces PLACE, a cutting-edge framework that enhances cross-modal alignment at the level of pathological observations for learning medical visual representation from image-report joint-training. This is achieved by a novel cross-modal pathological-level alignment module without the need for additional human annotations. Additionally, we develop a new proxy task, termed Cross-modal Correlation Exploration, which enables our model to capture more critical fine-grained details. These reasonable and carefully designed modules within PLACE allow it to learn a more generalizable and fine-grained medical visual representation. Experimental evaluations across multiple downstream tasks demonstrate the superiority of PLACE over previous state-of-the-art models, offering a robust solution to automatically align pathological information between images and reports without reliance on external human annotations.

\section*{{Limitations}}

{Our objective is to enhance medical visual representation learning by improving pathological-level cross-modal alignment and fine-grained representation learning, allowing seamless adaptation to different datasets or domains without requiring additional annotations such as bounding boxes or disease labels. Therefore, an explicit mechanism for distinguishing the abnormalities in the observation alignment is not present; instead, the reliance is on the proposed PCMA module and T-POR to implicitly navigate through pathological conditions. Thus there might be inconsistencies in aligning pathological conditions due to imperfections in the solution, e.g., normal V-POR to abnormal T-POR, even though the PCMA module at least enforces consistency by ensuring that pathological observations related to a specific anatomical region align correctly within samples. %For example, the V-POR for the lung should be closer to the T-POR for the lung rather than to regions like the heart. 
Furthermore, the number of parameters in the CCE module increases inversely with the fourth power of the patch size, potentially restricting applications that necessitate large image resolution using a small patch size. Our future work will focus on addressing such limitations without requiring additional annotations.}

\section*{Acknowledgments}
This work was supported in part by the UK Engineering and Physical Sciences Research Council (EPSRC) through a Turing AI Fellowship (grant no. EP/V020579/1, EP/V020579/2). 

% \appendices
% \input{sec/X_suppl}

\section*{References}

\bibliographystyle{IEEEtran}
\bibliography{refs}

\end{document}